# Thumbs Up or Thumbs Down? Semantic Orientation Applied to Unsupervised Classification of Reviews

**Peter D. Turney**
Institute for Information Technology
National Research Council of Canada
Ottawa, Ontario, Canada, K1A 0R6
`peter.turney@nrc.ca`

## Abstract

This paper presents a simple unsupervised learning algorithm for classifying reviews as *recommended* (thumbs up) or *not recommended* (thumbs down). The classification of a review is predicted by the average *semantic orientation* of the phrases in the review that contain adjectives or adverbs. A phrase has a positive semantic orientation when it has good associations (e.g., "subtle nuances") and a negative semantic orientation when it has bad associations (e.g., "very cavalier"). In this paper, the semantic orientation of a phrase is calculated as the mutual information between the given phrase and the word "excellent" minus the mutual information between the given phrase and the word "poor". A review is classified as recommended if the average semantic orientation of its phrases is positive. The algorithm achieves an average accuracy of 74% when evaluated on 410 reviews from Epinions, sampled from four different domains (reviews of automobiles, banks, movies, and travel destinations). The accuracy ranges from 84% for automobile reviews to 66% for movie reviews.

## 1  Introduction

If you are considering a vacation in Akumal, Mexico, you might go to a search engine and enter the query "Akumal travel review". However, in this case, Google[1] reports about 5,000 matches. It would be useful to know what fraction of these matches recommend Akumal as a travel destination. With an algorithm for automatically classifying a review as "thumbs up" or "thumbs down", it would be possible for a search engine to report such summary statistics. This is the motivation for the research described here. Other potential applications include recognizing "flames" (abusive newsgroup messages) (Spertus, 1997) and developing new kinds of search tools (Hearst, 1992).

In this paper, I present a simple unsupervised learning algorithm for classifying a review as *recommended* or *not recommended*. The algorithm takes a written review as input and produces a classification as output. The first step is to use a part-of-speech tagger to identify phrases in the input text that contain adjectives or adverbs (Brill, 1994). The second step is to estimate the *semantic orientation* of each extracted phrase (Hatzivassiloglou & McKeown, 1997). A phrase has a positive semantic orientation when it has good associations (e.g., "romantic ambience") and a negative semantic orientation when it has bad associations (e.g., "horrific events"). The third step is to assign the given review to a class, *recommended* or *not recommended*, based on the average semantic orientation of the phrases extracted from the review. If the average is positive, the prediction is that the review recommends the item it discusses. Otherwise, the prediction is that the item is not recommended.

The PMI-IR algorithm is employed to estimate the semantic orientation of a phrase (Turney, 2001). PMI-IR uses Pointwise Mutual Information (PMI) and Information Retrieval (IR) to measure the similarity of pairs of words or phrases. The se-mantic orientation of a given phrase is calculated by comparing its similarity to a positive reference word ("excellent") with its similarity to a negative reference word ("poor"). More specifically, a phrase is assigned a numerical rating by taking the mutual information between the given phrase and the word "excellent" and subtracting the mutual information between the given phrase and the word "poor". In addition to determining the direction of the phrase's semantic orientation (positive or negative, based on the sign of the rating), this numerical rating also indicates the strength of the semantic orientation (based on the magnitude of the number). The algorithm is presented in Section 2.

---
[1] http://www.google.com

Hatzivassiloglou and McKeown (1997) have also developed an algorithm for predicting semantic orientation. Their algorithm performs well, but it is designed for isolated adjectives, rather than phrases containing adjectives or adverbs. This is discussed in more detail in Section 3, along with other related work.

The classification algorithm is evaluated on 410 reviews from Epinions[2], randomly sampled from four different domains: reviews of automobiles, banks, movies, and travel destinations. Reviews at Epinions are not written by professional writers; any person with a Web browser can become a member of Epinions and contribute a review. Each of these 410 reviews was written by a different author. Of these reviews, 170 are *not recommended* and the remaining 240 are *recommended* (these classifications are given by the authors). Always guessing the majority class would yield an accuracy of 59%. The algorithm achieves an average accuracy of 74%, ranging from 84% for automobile reviews to 66% for movie reviews. The experimental results are given in Section 4.

The interpretation of the experimental results, the limitations of this work, and future work are discussed in Section 5. Potential applications are outlined in Section 6. Finally, conclusions are presented in Section 7.

## 2 Classifying Reviews

The first step of the algorithm is to extract phrases containing adjectives or adverbs. Past work has demonstrated that adjectives are good indicators of subjective, evaluative sentences (Hatzivassiloglou & Wiebe, 2000; Wiebe, 2000; Wiebe et al., 2001). However, although an isolated adjective may indicate subjectivity, there may be insufficient context to determine semantic orientation. For example, the adjective "unpredictable" may have a negative orientation in an automotive review, in a phrase such as "unpredictable steering", but it could have a positive orientation in a movie review, in a phrase such as "unpredictable plot". Therefore the algorithm extracts two consecutive words, where one member of the pair is an adjective or an adverb and the second provides context.

First a part-of-speech tagger is applied to the review (Brill, 1994).[3] Two consecutive words are extracted from the review if their tags conform to any of the patterns in Table 1. The JJ tags indicate adjectives, the NN tags are nouns, the RB tags are adverbs, and the VB tags are verbs.[4] The second pattern, for example, means that two consecutive words are extracted if the first word is an adverb and the second word is an adjective, but the third word (which is not extracted) cannot be a noun. NNP and NNPS (singular and plural proper nouns) are avoided, so that the names of the items in the review cannot influence the classification.

Table 1. Patterns of tags for extracting two-word phrases from reviews.

| | First Word | Second Word | Third Word (Not Extracted) |
|---|---|---|---|
| 1. | JJ | NN or NNS | anything |
| 2. | RB, RBR, or RBS | JJ | not NN nor NNS |
| 3. | JJ | JJ | not NN nor NNS |
| 4. | NN or NNS | JJ | not NN nor NNS |
| 5. | RB, RBR, or RBS | VB, VBD, VBN, or VBG | anything |

The second step is to estimate the semantic orientation of the extracted phrases, using the PMI-IR algorithm. This algorithm uses mutual information as a measure of the strength of semantic association between two words (Church & Hanks, 1989). PMI-IR has been empirically evaluated using 80 synonym test questions from the Test of English as a Foreign Language (TOEFL), obtaining a score of 74% (Turney, 2001). For comparison, Latent Semantic Analysis (LSA), another statistical measure of word association, attains a score of 64% on the same 80 TOEFL questions (Landauer & Dumais, 1997).

---

[2] http://www.epinions.com

[3] http://www.cs.jhu.edu/~brill/RBT1_14.tar.Z

[4] See Santorini (1995) for a complete description of the tags.

The Pointwise Mutual Information (PMI) between two words, $word_1$ and $word_2$, is defined as follows (Church & Hanks, 1989):

$$\text{PMI}(word_1, word_2) = \log_2 \left[ \frac{p(word_1 \,\&\, word_2)}{p(word_1)\, p(word_2)} \right] \quad (1)$$

Here, $p(word_1 \,\&\, word_2)$ is the probability that $word_1$ and $word_2$ co-occur. If the words are statistically independent, then the probability that they co-occur is given by the product $p(word_1)\, p(word_2)$. The ratio between $p(word_1 \,\&\, word_2)$ and $p(word_1)\, p(word_2)$ is thus a measure of the degree of statistical dependence between the words. The log of this ratio is the amount of information that we acquire about the presence of one of the words when we observe the other.

The Semantic Orientation (SO) of a phrase, *phrase*, is calculated here as follows:

$$\text{SO}(phrase) = \text{PMI}(phrase, \text{``excellent''}) - \text{PMI}(phrase, \text{``poor''}) \quad (2)$$

The reference words "excellent" and "poor" were chosen because, in the five star review rating system, it is common to define one star as "poor" and five stars as "excellent". SO is positive when *phrase* is more strongly associated with "excellent" and negative when *phrase* is more strongly associated with "poor".

PMI-IR estimates PMI by issuing queries to a search engine (hence the IR in PMI-IR) and noting the number of hits (matching documents). The following experiments use the AltaVista Advanced Search engine[5], which indexes approximately 350 million web pages (counting only those pages that are in English). I chose AltaVista because it has a NEAR operator. The AltaVista NEAR operator constrains the search to documents that contain the words within ten words of one another, in either order. Previous work has shown that NEAR performs better than AND when measuring the strength of semantic association between words (Turney, 2001).

Let hits(*query*) be the number of hits returned, given the query *query*. The following estimate of SO can be derived from equations (1) and (2) with some minor algebraic manipulation, if co-occurrence is interpreted as NEAR:

$$\text{SO}(phrase) =$$
$$\log_2 \left[ \frac{\text{hits}(phrase\ \text{NEAR ``excellent''})\ \text{hits}(\text{``poor''})}{\text{hits}(phrase\ \text{NEAR ``poor''})\ \text{hits}(\text{``excellent''})} \right] \quad (3)$$

Equation (3) is a log-odds ratio (Agresti, 1996). To avoid division by zero, I added 0.01 to the hits. I also skipped *phrase* when both hits(*phrase* NEAR "excellent") and hits(*phrase* NEAR "poor") were (simultaneously) less than four. These numbers (0.01 and 4) were arbitrarily chosen. To eliminate any possible influence from the testing data, I added "AND (NOT host:epinions)" to every query, which tells AltaVista not to include the Epinions Web site in its searches.

The third step is to calculate the average semantic orientation of the phrases in the given review and classify the review as *recommended* if the average is positive and otherwise *not recommended*.

Table 2 shows an example for a *recommended* review and Table 3 shows an example for a *not recommended* review. Both are reviews of the Bank of America. Both are in the collection of 410 reviews from Epinions that are used in the experiments in Section 4.

Table 2. An example of the processing of a review that the author has classified as *recommended*.[6]

| Extracted Phrase | Part-of-Speech Tags | Semantic Orientation |
|---|---|---|
| online experience | JJ NN | 2.253 |
| low fees | JJ NNS | 0.333 |
| local branch | JJ NN | 0.421 |
| small part | JJ NN | 0.053 |
| online service | JJ NN | 2.780 |
| printable version | JJ NN | -0.705 |
| direct deposit | JJ NN | 1.288 |
| well other | RB JJ | 0.237 |
| inconveniently located | RB VBN | -1.541 |
| other bank | JJ NN | -0.850 |
| true service | JJ NN | -0.732 |
| Average Semantic Orientation | | 0.322 |

---

[5] http://www.altavista.com/sites/search/adv

[6] The semantic orientation in the following tables is calculated using the natural logarithm (base *e*), rather than base 2. The natural log is more common in the literature on log-odds ratio. Since all logs are equivalent up to a constant factor, it makes no difference for the algorithm.

Table 3. An example of the processing of a review that the author has classified as *not recommended*.

| Extracted Phrase | Part-of-Speech Tags | Semantic Orientation |
|---|---|---|
| little difference | JJ NN | -1.615 |
| clever tricks | JJ NNS | -0.040 |
| programs such | NNS JJ | 0.117 |
| possible moment | JJ NN | -0.668 |
| unethical practices | JJ NNS | -8.484 |
| low funds | JJ NNS | -6.843 |
| old man | JJ NN | -2.566 |
| other problems | JJ NNS | -2.748 |
| probably wondering | RB VBG | -1.830 |
| virtual monopoly | JJ NN | -2.050 |
| other bank | JJ NN | -0.850 |
| extra day | JJ NN | -0.286 |
| direct deposits | JJ NNS | 5.771 |
| online web | JJ NN | 1.936 |
| cool thing | JJ NN | 0.395 |
| very handy | RB JJ | 1.349 |
| lesser evil | RBR JJ | -2.288 |
| Average Semantic Orientation | | -1.218 |

## 3 Related Work

This work is most closely related to Hatzivassiloglou and McKeown's (1997) work on predicting the semantic orientation of adjectives. They note that there are linguistic constraints on the semantic orientations of adjectives in conjunctions. As an example, they present the following three sentences (Hatzivassiloglou & McKeown, 1997):

1. The tax proposal was simple and well-received by the public.

2. The tax proposal was simplistic but well-received by the public.

3. (*) The tax proposal was simplistic and well-received by the public.

The third sentence is incorrect, because we use "and" with adjectives that have the same semantic orientation ("simple" and "well-received" are both positive), but we use "but" with adjectives that have different semantic orientations ("simplistic" is negative).

Hatzivassiloglou and McKeown (1997) use a four-step supervised learning algorithm to infer the semantic orientation of adjectives from constraints on conjunctions:

1. All conjunctions of adjectives are extracted from the given corpus.

2. A supervised learning algorithm combines multiple sources of evidence to label pairs of adjectives as having the *same semantic orientation* or *different semantic orientations*. The result is a graph where the nodes are adjectives and links indicate sameness or difference of semantic orientation.

3. A clustering algorithm processes the graph structure to produce two subsets of adjectives, such that links across the two subsets are mainly different-orientation links, and links inside a subset are mainly same-orientation links.

4. Since it is known that positive adjectives tend to be used more frequently than negative adjectives, the cluster with the higher average frequency is classified as having positive semantic orientation.

This algorithm classifies adjectives with accuracies ranging from 78% to 92%, depending on the amount of training data that is available. The algorithm can go beyond a binary *positive-negative* distinction, because the clustering algorithm (step 3 above) can produce a "goodness-of-fit" measure that indicates how well an adjective fits in its assigned cluster.

Although they do not consider the task of classifying reviews, it seems their algorithm could be plugged into the classification algorithm presented in Section 2, where it would replace PMI-IR and equation (3) in the second step. However, PMI-IR is conceptually simpler, easier to implement, and it can handle phrases and adverbs, in addition to isolated adjectives.

As far as I know, the only prior published work on the task of classifying reviews as thumbs up or down is Tong's (2001) system for generating *sentiment timelines*. This system tracks online discussions about movies and displays a plot of the number of positive sentiment and negative sentiment messages over time. Messages are classified by looking for specific phrases that indicate the sentiment of the author towards the movie (e.g., "great acting", "wonderful visuals", "terrible score", "uneven editing"). Each phrase must be manually added to a special lexicon and manually tagged as indicating positive or negative sentiment. The lexicon is specific to the domain (e.g., movies) and must be built anew for each new domain. The company Mindfuleye[7] offers a technology called Lexant™ that appears similar to Tong's (2001) system.

Other related work is concerned with determining subjectivity (Hatzivassiloglou & Wiebe, 2000; Wiebe, 2000; Wiebe et al., 2001). The task is to distinguish sentences that present opinions and evaluations from sentences that objectively present factual information (Wiebe, 2000). Wiebe *et al.* (2001) list a variety of potential applications for automated subjectivity tagging, such as recognizing "flames" (Spertus, 1997), classifying email, recognizing speaker role in radio broadcasts, and mining reviews. In several of these applications, the first step is to recognize that the text is subjective and then the natural second step is to determine the semantic orientation of the subjective text. For example, a flame detector cannot merely detect that a newsgroup message is subjective, it must further detect that the message has a negative semantic orientation; otherwise a message of praise could be classified as a flame.

Hearst (1992) observes that most search engines focus on finding documents on a given *topic*, but do not allow the user to specify the *directionality* of the documents (e.g., is the author in favor of, neutral, or opposed to the event or item discussed in the document?). The directionality of a document is determined by its deep argumentative structure, rather than a shallow analysis of its adjectives. Sentences are interpreted metaphorically in terms of agents exerting force, resisting force, and overcoming resistance. It seems likely that there could be some benefit to combining shallow and deep analysis of the text.

## 4 Experiments

Table 4 describes the 410 reviews from Epinions that were used in the experiments. 170 (41%) of the reviews are *not recommended* and the remaining 240 (59%) are *recommended*. Always guessing the majority class would yield an accuracy of 59%. The third column shows the average number of phrases that were extracted from the reviews.

Table 5 shows the experimental results. Except for the travel reviews, there is surprisingly little variation in the accuracy within a domain. In addition to *recommended* and *not recommended*, Epinions reviews are classified using the five star rating system. The third column shows the correlation between the average semantic orientation and the number of stars assigned by the author of the review. The results show a strong positive correlation between the average semantic orientation and the author's rating out of five stars.

Table 4. A summary of the corpus of reviews.

| Domain of Review | Number of Reviews | Average Phrases per Review |
|---|---|---|
| Automobiles | 75 | 20.87 |
|   Honda Accord | 37 | 18.78 |
|   Volkswagen Jetta | 38 | 22.89 |
| Banks | 120 | 18.52 |
|   Bank of America | 60 | 22.02 |
|   Washington Mutual | 60 | 15.02 |
| Movies | 120 | 29.13 |
|   The Matrix | 60 | 19.08 |
|   Pearl Harbor | 60 | 39.17 |
| Travel Destinations | 95 | 35.54 |
|   Cancun | 59 | 30.02 |
|   Puerto Vallarta | 36 | 44.58 |
| All | 410 | 26.00 |

Table 5. The accuracy of the classification and the correlation of the semantic orientation with the star rating.

| Domain of Review | Accuracy | Correlation |
|---|---|---|
| Automobiles | 84.00 % | 0.4618 |
|   Honda Accord | 83.78 % | 0.2721 |
|   Volkswagen Jetta | 84.21 % | 0.6299 |
| Banks | 80.00 % | 0.6167 |
|   Bank of America | 78.33 % | 0.6423 |
|   Washington Mutual | 81.67 % | 0.5896 |
| Movies | 65.83 % | 0.3608 |
|   The Matrix | 66.67 % | 0.3811 |
|   Pearl Harbor | 65.00 % | 0.2907 |
| Travel Destinations | 70.53 % | 0.4155 |
|   Cancun | 64.41 % | 0.4194 |
|   Puerto Vallarta | 80.56 % | 0.1447 |
| All | 74.39 % | 0.5174 |

## 5 Discussion of Results

A natural question, given the preceding results, is what makes movie reviews hard to classify? Table 6 shows that classification by the average SO tends to err on the side of guessing that a review is *not recommended*, when it is actually *recommended*. This suggests the hypothesis that a good movie will often contain unpleasant scenes (e.g., violence, death, mayhem), and a *recommended* movie re-view may thus have its average semantic orientation reduced if it contains descriptions of these unpleasant scenes. However, if we add a constant value to the average SO of the movie reviews, to compensate for this bias, the accuracy does not improve. This suggests that, just as positive reviews mention unpleasant things, so negative reviews often mention pleasant scenes.

---
[7] http://www.mindfuleye.com/

Table 6. The confusion matrix for movie classifications.

| Average Semantic Orientation | Author's Classification | | Sum |
|---|---|---|---|
| | Thumbs Up | Thumbs Down | |
| Positive | 28.33 % | 12.50 % | 40.83 % |
| Negative | 21.67 % | 37.50 % | 59.17 % |
| Sum | 50.00 % | 50.00 % | 100.00 % |

Table 7 shows some examples that lend support to this hypothesis. For example, the phrase "more evil" does have negative connotations, thus an SO of -4.384 is appropriate, but an evil character does not make a bad movie. The difficulty with movie reviews is that there are two aspects to a movie, the events and actors in the movie (the elements of the movie), and the style and art of the movie (the movie as a gestalt; a unified whole). This is likely also the explanation for the lower accuracy of the Cancun reviews: good beaches do not necessarily add up to a good vacation. On the other hand, good automotive parts usually do add up to a good automobile and good banking services add up to a good bank. It is not clear how to address this issue. Future work might look at whether it is possible to tag sentences as discussing *elements* or *wholes*.

Another area for future work is to empirically compare PMI-IR and the algorithm of Hatzivassiloglou and McKeown (1997). Although their algorithm does not readily extend to two-word phrases, I have not yet demonstrated that two-word phrases are necessary for accurate classification of reviews. On the other hand, it would be interesting to evaluate PMI-IR on the collection of 1,336 hand-labeled adjectives that were used in the experiments of Hatzivassiloglou and McKeown (1997). A related question for future work is the relationship of accuracy of the estimation of semantic orientation at the level of individual phrases to accuracy of review classification. Since the review classification is based on an average, it might be quite resistant to noise in the SO estimate for individual phrases.

But it is possible that a better SO estimator could produce significantly better classifications.

Table 7. Sample phrases from misclassified reviews.

| Movie: | The Matrix |
|---|---|
| Author's Rating: | *recommended* (5 stars) |
| Average SO: | -0.219 (*not recommended*) |
| Sample Phrase: | more evil   [RBR JJ] |
| SO of Sample Phrase: | -4.384 |
| Context of Sample Phrase: | The slow, methodical way he spoke. I loved it! It made him seem more arrogant and even **more evil**. |
| Movie: | Pearl Harbor |
| Author's Rating: | *recommended* (5 stars) |
| Average SO: | -0.378 (*not recommended*) |
| Sample Phrase: | sick feeling   [JJ NN] |
| SO of Sample Phrase: | -8.308 |
| Context of Sample Phrase: | During this period I had a **sick feeling**, knowing what was coming, knowing what was part of our history. |
| Movie: | The Matrix |
| Author's Rating: | *not recommended* (2 stars) |
| Average SO: | 0.177 (*recommended*) |
| Sample Phrase: | very talented   [RB JJ] |
| SO of Sample Phrase: | 1.992 |
| Context of Sample Phrase: | Well as usual Keanu Reeves is nothing special, but surprisingly, the **very talented** Laurence Fishbourne is not so good either, I was surprised. |
| Movie: | Pearl Harbor |
| Author's Rating: | *not recommended* (3 stars) |
| Average SO: | 0.015 (*recommended*) |
| Sample Phrase: | blue skies   [JJ NNS] |
| SO of Sample Phrase: | 1.263 |
| Context of Sample Phrase: | Anyone who saw the trailer in the theater over the course of the last year will never forget the images of Japanese war planes swooping out of the **blue skies**, flying past the children playing baseball, or the truly remarkable shot of a bomb falling from an enemy plane into the deck of the USS Arizona. |

Equation (3) is a very simple estimator of semantic orientation. It might benefit from more sophisticated statistical analysis (Agresti, 1996). One possibility is to apply a statistical significance test to each estimated SO. There is a large statistical literature on the log-odds ratio, which might lead to improved results on this task.

This paper has focused on unsupervised classification, but average semantic orientation could be supplemented by other features, in a supervised classification system. The other features could be based on the presence or absence of specific words, as is common in most text classification work. This could yield higher accuracies, but the intent here was to study this one feature in isolation, to simplify the analysis, before combining it with other features.

Table 5 shows a high correlation between the average semantic orientation and the star rating of a review. I plan to experiment with ordinal classification of reviews in the five star rating system, using the algorithm of Frank and Hall (2001). For ordinal classification, the average semantic orientation would be supplemented with other features in a supervised classification system.

A limitation of PMI-IR is the time required to send queries to AltaVista. Inspection of Equation (3) shows that it takes four queries to calculate the semantic orientation of a phrase. However, I cached all query results, and since there is no need to recalculate hits("poor") and hits("excellent") for every phrase, each phrase requires an average of slightly less than two queries. As a courtesy to AltaVista, I used a five second delay between queries.[8] The 410 reviews yielded 10,658 phrases, so the total time required to process the corpus was roughly 106,580 seconds, or about 30 hours.

This might appear to be a significant limitation, but extrapolation of current trends in computer memory capacity suggests that, in about ten years, the average desktop computer will be able to easily store and search AltaVista's 350 million Web pages. This will reduce the processing time to less than one second per review.

## 6 Applications

There are a variety of potential applications for automated review rating. As mentioned in the introduction, one application is to provide summary statistics for search engines. Given the query "Akumal travel review", a search engine could report, "There are 5,000 hits, of which 80% are thumbs up and 20% are thumbs down." The search results could be sorted by average semantic orientation, so that the user could easily sample the most extreme reviews. Similarly, a search engine could allow the user to specify the topic and the rating of the desired reviews (Hearst, 1992).

Preliminary experiments indicate that semantic orientation is also useful for summarization of reviews. A positive review could be summarized by picking out the sentence with the highest positive semantic orientation and a negative review could be summarized by extracting the sentence with the lowest negative semantic orientation.

Epinions asks its reviewers to provide a short description of *pros* and *cons* for the reviewed item. A pro/con summarizer could be evaluated by measuring the overlap between the reviewer's pros and cons and the phrases in the review that have the most extreme semantic orientation.

Another potential application is filtering "flames" for newsgroups (Spertus, 1997). There could be a threshold, such that a newsgroup message is held for verification by the human moderator when the semantic orientation of a phrase drops below the threshold. A related use might be a tool for helping academic referees when reviewing journal and conference papers. Ideally, referees are unbiased and objective, but sometimes their criticism can be unintentionally harsh. It might be possible to highlight passages in a draft referee's report, where the choice of words should be modified towards a more neutral tone.

Tong's (2001) system for detecting and tracking opinions in on-line discussions could benefit from the use of a learning algorithm, instead of (or in addition to) a hand-built lexicon. With automated review rating (opinion rating), advertisers could track advertising campaigns, politicians could track public opinion, reporters could track public response to current events, stock traders could track financial opinions, and trend analyzers could track entertainment and technology trends.

## 7 Conclusions

This paper introduces a simple unsupervised learning algorithm for rating a review as thumbs up or down. The algorithm has three steps: (1) extract phrases containing adjectives or adverbs, (2) estimate the semantic orientation of each phrase, and (3) classify the review based on the average semantic orientation of the phrases. The core of the algorithm is the second step, which uses PMI-IR to calculate semantic orientation (Turney, 2001).

---

[8] This line of research depends on the good will of the major search engines. For a discussion of the ethics of Web robots, see http://www.robotstxt.org/wc/robots.html. For query robots, the proposed extended standard for robot exclusion would be useful. See http://www.conman.org/people/spc/robots2.html.

In experiments with 410 reviews from Epinions, the algorithm attains an average accuracy of 74%. It appears that movie reviews are difficult to classify, because the whole is not necessarily the sum of the parts; thus the accuracy on movie reviews is about 66%. On the other hand, for banks and automobiles, it seems that the whole is the sum of the parts, and the accuracy is 80% to 84%. Travel reviews are an intermediate case.

Previous work on determining the semantic orientation of adjectives has used a complex algorithm that does not readily extend beyond isolated adjectives to adverbs or longer phrases (Hatzivassiloglou and McKeown, 1997). The simplicity of PMI-IR may encourage further work with semantic orientation.

The limitations of this work include the time required for queries and, for some applications, the level of accuracy that was achieved. The former difficulty will be eliminated by progress in hardware. The latter difficulty might be addressed by using semantic orientation combined with other features in a supervised classification algorithm.

## Acknowledgements

Thanks to Joel Martin and Michael Littman for helpful comments.